\documentclass[10pt,twocolumn,letterpaper]{article}

\usepackage{iccv}
\usepackage{times}
\usepackage{epsfig}
\usepackage{graphicx}
\usepackage{amsmath}
\usepackage{amssymb}
\usepackage{booktabs}

\usepackage{multirow}

\usepackage{algorithm}
\usepackage{algorithmic}

\usepackage{xcolor}

\usepackage[accsupp]{axessibility}  

\usepackage[pagebackref=true,breaklinks=true,letterpaper=true,colorlinks,bookmarks=false]{hyperref}

\iccvfinalcopy 

\def\iccvPaperID{5673} 

\begin{document}

\title{Efficient Model Personalization in Federated Learning \\ via Client-Specific Prompt Generation}

\author{Fu-En Yang$^{1,2}$ \qquad Chien-Yi Wang$^{2}$ \qquad Yu-Chiang Frank Wang$^{1, 2}$ \\
$^{1}$National Taiwan University \qquad $^{2}$NVIDIA\\
{\tt\small \{f07942077, ycwang\}@ntu.edu.tw, chienyiw@nvidia.com}
}


\maketitle

\begin{abstract}
Federated learning (FL) emerges as a decentralized learning framework which trains models from multiple distributed clients without sharing their data to preserve privacy. Recently, large-scale pre-trained models (e.g., Vision Transformer) have shown a strong capability of deriving robust representations. However, the data heterogeneity among clients, the limited computation resources, and the communication bandwidth restrict the deployment of large-scale models in FL frameworks. To leverage robust representations from large-scale models while enabling efficient model personalization for heterogeneous clients, we propose a novel personalized FL framework of client-specific Prompt Generation (pFedPG), which learns to deploy a personalized prompt generator at the server for producing client-specific visual prompts that efficiently adapts frozen backbones to local data distributions. Our proposed framework jointly optimizes the stages of personalized prompt adaptation locally and personalized prompt generation globally. The former aims to train visual prompts that adapt foundation models to each client, while the latter observes local optimization directions to generate personalized prompts for all clients. Through extensive experiments on benchmark datasets, we show that our pFedPG is favorable against state-of-the-art personalized FL methods under various types of data heterogeneity, allowing computation and communication efficient model personalization.

\end{abstract}

\section{Introduction}
\label{sec:intro}

With access to web-scale training data (\textit{e.g.,} LAION-5B~\cite{schuhmann2022laion}), deep learning has demonstrated remarkable achievements across computer vision~\cite{he2020momentum,he2022masked,radford2021learning} and natural language understanding~\cite{devlin2018bert,zhang2022opt,brown2020language}.
However, in real-world scenarios, user data is typically scattered across various domains, such as hospital sites or edge devices. Due to increasing risks of privacy breaches and stricter privacy protection regulations~\cite{custers2019eu}, centralized learning schemes are not preferable.
With the aim of collaboratively training models without exposing users' private data, Federated learning (FL) has emerged as a prominent distributed learning framework and has garnered growing research interest. This privacy-preserving learning paradigm has been widely adopted in applications like medical image diagnosis~\cite{chen2021personalized}, face recognition~\cite{liu2022fedfr}, and person re-identification~\cite{zhuang2020performance}.

\begin{figure}[t!]
	\centering
	\includegraphics[width=0.40\textwidth]{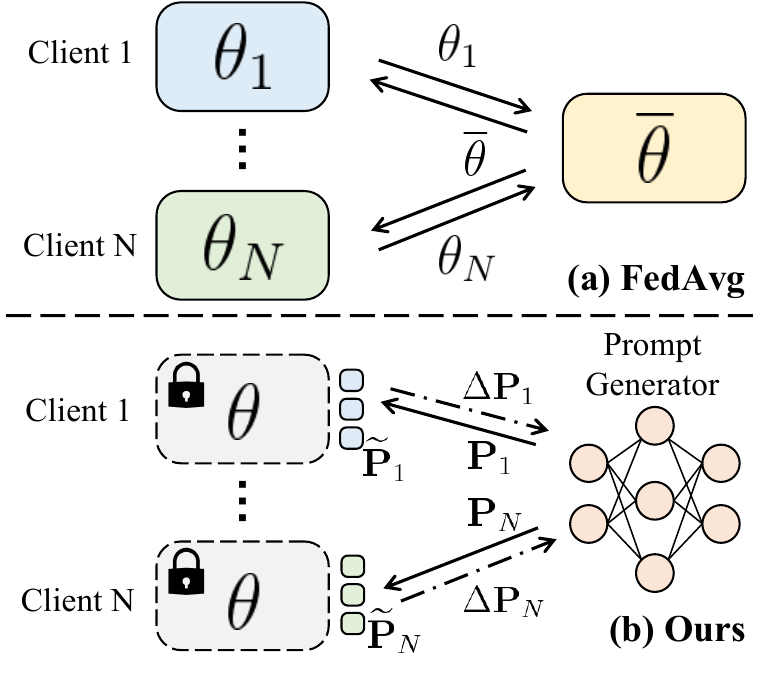}
 	\vspace{0mm}
    \caption{Comparison between (a) FedAvg and (b) our approach. Instead of updating and transporting entire models $\theta$, our FL method learns to generate personalized prompts $\textbf{P}$ by implicitly observing local optimization directions $\Delta \textbf{P}=\widetilde{\textbf{P}}-\textbf{P}$ for efficient model personalization on top of frozen foundation models.
    }
    \vspace{0mm}
	\label{fig:intro}
\end{figure}

Without the need of data sharing among clients, the mainstream FL approach of FedAvg~\cite{mcmahan2017communication} learns a global model by averaging model parameters trained on clients' private data.
However, data distributed in each client might be \emph{heterogeneous} in terms of \emph{domain discrepancy}~\cite{li2021fedbn} or \emph{imbalanced class distribution}~\cite{li2021model}. Sharing a global model across heterogeneous data clients is prone to highly deviate from their local distribution, leading to severe performance degradation~\cite{shamsian2021personalized,ma2022layer}. 
Previous FL works~\cite{li2020federated,li2021model} propose types of constraints (\textit{e.g.,} $L_2$~\cite{li2020federated} or contrastive regularization~\cite{li2021model}) to prevent the local training to be divergent from each other.
To better handle the inevitable data heterogeneity across clients, personalized federated learning (pFL) methods~\cite{shamsian2021personalized,ma2022layer,chen2021bridging,zhang2022fedala,shen2022cd2} are instead proposed to allow each client to train a personalized model that adapts to their own data distribution.
For example, pFedHN~\cite{shamsian2021personalized} introduces a hypernetwork at the server to directly generate model parameters for each client, whereas pFedLA~\cite{ma2022layer} learns a layer-wise model aggregation policy to assign different weights for personalized model aggregation.
While the above pFL approaches are desirable for handling heterogeneous data, they are typically restricted to small backbone architectures (\textit{e.g.,} LeNet~\cite{lecun1998gradient}) due to the high complexity of outputting model parameters~\cite{shamsian2021personalized} or aggregation weights~\cite{ma2022layer} for large-scale models. Consequently, the capability of derived features is limited, leading to a lack of performance improvement and training instability.

Recently, training from large foundation models~\cite{bommasani2021opportunities} for downstream tasks has become a prominent paradigm in centralized learning. To leverage the strong representations derived by foundation models for alleviating data heterogeneity, ViT-FL~\cite{qu2022rethinking} incorporates pre-trained Vision Transformer (ViT)~\cite{dosovitskiy2020vit} into standard FL algorithms (\textit{e.g.,} FedAvg~\cite{mcmahan2017communication}) and shows improved robustness and stability on heterogeneously distributed data.
However, the use of large pre-trained models for all clients in existing FL algorithms can cause extensive computational and communication burdens, as these methods require transporting entire model parameters between clients and the server. Additionally, overfitting issues might occur when large-scale models are trained with relatively limited client data.

For efficiently tuning large-scale models, prompt learning~\cite{jia2022visual,zhou2022cocoop,zhou2022coop} provides a flexible way to adapt pre-trained models to downstream tasks by solely training the additional inserted trainable parameters (\textit{i.e.,} prompts). For instance, VPT~\cite{jia2022visual} treats prompts as task-specific parameters and prepends them to the input tokens of a pre-trained ViT. In this way, prompts could be optimized to capture task-specific information while instructing a frozen model to perform tasks of interest.
However, a straightforward way to adopt prompt learning into FL, \textit{i.e.,} simply averaging prompts learned from all clients, cannot address data heterogeneity among clients effectively and often leads to unsatisfactory performance (as evident in Tables~\ref{tab:domain}-\ref{tab:medical}). Therefore, there is a crucial challenge to develop new FL methods that can leverage prompt learning effectively while handling data heterogeneity among clients.

In this paper, we aim at achieving efficient model personalization among clients with data heterogeneity. As depicted in Fig.~\ref{fig:intro}, different from conventional FL methods (\textit{e.g.,} FedAvg~\cite{mcmahan2017communication}) that updates and transports entire model parameters, we propose a novel personalized FL scheme of \emph{client-specific Prompt Generation (pFedPG)} that exploits underlying client-specific characteristics to produce personalized prompts for each client, which enables efficient adaptation to local data distribution.
To be more precise, each client trains the client-specific prompts to instruct a model to perform recognition tasks on the target client using its private data. As the local training is not required to update entire large models, the computation overload could be minimized while the possible overfitting issues are mitigated accordingly.
On the other hand, we employ a personalized prompt generation module on the server side, which is learned to obtain the underlying optimization directions among clients. With such client characteristics implicitly observed, we are capable of producing personalized prompts to facilitate efficient adaptation for each client with heterogeneous data distribution. 
By iteratively training the above two stages in a mutually beneficial manner, we are capable of achieving effective yet efficient model personalization on top of the robust representations derived from large-scale foundation models.

We now summarize the contributions of this work below:
\begin{itemize}

\item We propose a personalized FL framework of client-specific Prompt Generation (pFedPG), which alternates between \textit{personalized prompt generation} and \textit{personalized prompt adaptation} to enable efficient model personalization under heterogeneous data.


\item We design a client-specific prompt generator at the server, which effectively exploits personalized optimization directions and produces client-specific prompts for updating each client model.


\item Evaluations on several benchmark datasets in domain discrepancy and imbalanced class distribution verify that our method performs favorably against existing personalized FL approaches and exhibits sufficient training efficiency. 

\end{itemize}

\section{Related Works}
\label{sec:related}

\paragraph{Federated Learning (FL)}
Federated Learning is a learning framework in machine learning with the goal of training models from distributed data sources while protecting data privacy. The most widely recognized approach for federated learning is FedAvg~\cite{mcmahan2017communication}, which partitions the learning process into local training and global averaging. 
However, data distributed in real-world scenarios are typically non-IID, indicating the presence of domain discrepancy or imbalanced class distribution among clients. Directly averaging models trained on heterogeneous data can lead to severe performance degradation and training instability.
To address this challenge, several methods~\cite{li2020federated,karimireddy2020scaffold,li2021model,sun2021partialfed,tan2022fedproto,zhang2022fine,mendieta2022local} have been proposed to regularize local training in FedAvg~\cite{mcmahan2017communication}. For instance, FedProx~\cite{li2020federated} and SCAFFOLD~\cite{karimireddy2020scaffold} restrict the local update to be consistent by $L_2$ distance over model weights and variance reduction technique over gradients, respectively. MOON~\cite{li2021model} applies a contrastive objective to regularize the optimization of local models, ensuring that they do not deviate significantly from the global model. 

\paragraph{Personalized Federated Learning (pFL)}
Instead of constructing a global model shared among all clients, personalized FL algorithms~\cite{li2021fedbn,fallah2020personalized,collins2021exploiting,li2021ditto,shamsian2021personalized,ma2022layer,chen2021bridging,oh2022fedbabu,zhang2022fedala,shen2022cd2,dai2022dispfl,shysheya2022fit,chen2023efficient} are proposed to address data heterogeneity issues by learning customized models at each client.
Several works~\cite{collins2021exploiting,oh2022fedbabu,chen2021bridging} achieve model personalization by only aggregating parts of a model (\textit{e.g.,} feature extractor) at the server while keeping or learning additional modules (\textit{e.g.,} classifier) locally. 
Per-FedAvg~\cite{fallah2020personalized} analogizes the local training and server aggregation processes as inner and outer loops optimization in model-agnostic meta-learning~\cite{finn2017model}, facilitating local model adaptation from the global model initialization.
PartialFed~\cite{sun2021partialfed} and FedALA~\cite{zhang2022fedala} derive customized models by adaptively aggregating the global and local models. Similarly, pFedLA~\cite{ma2022layer} learns a layer-wise aggregation policy to construct a personalized model by assigning larger weights to clients with higher similarities. 
Some recent works~\cite{dai2022dispfl,shysheya2022fit,chen2023efficient} achieve model personalization by either learning sparse models or applying adapter layers.
Instead of employing average-based aggregation at the server, pFedHN~\cite{shamsian2021personalized} directly generates model parameters for all clients. However, its applicability is limited to small and shallow models (\textit{e.g.,} LeNet~\cite{lecun1998gradient}) due to the high complexity of the model parameter space.

\begin{figure*}[t!]
	\centering
	\includegraphics[width=0.85\textwidth]{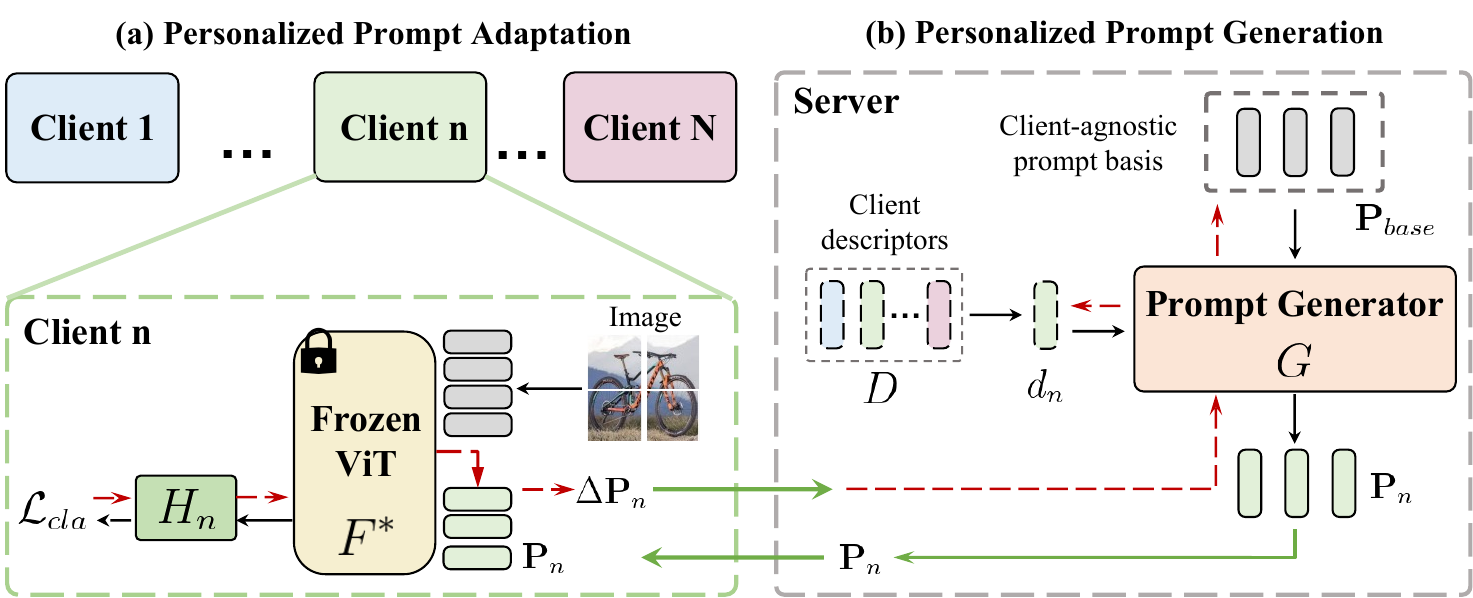}
 	\vspace{1mm}
    \caption{Overview of our client-specific Prompt Generation (pFedPG) framework. pFedPG learns a prompt generator $G$ together with client-agnostic prompt basis $\textbf{P}_{base}$ and a bank of client descriptors $D=\{d_n\}_{n=1}^N$ at the server. With local classification loss observed, both client-specific prompts $\mathtt{\textbf{P}}_n$ and local classification head $H_n$ are updated at each client $n$. We alternate between the stages of (a) \textit{personalized prompt adaptation} and (b) \textit{personalized prompt generation} to enable efficient personalization of foundation models like ViT.
    }
    \vspace{-2mm}
	\label{fig:archi}
\end{figure*}

\paragraph{Foundation Models and Prompt Learning}

Leveraging publicly available pre-trained foundation models~\cite{bommasani2021opportunities,dosovitskiy2020vit,he2020momentum,he2022masked,radford2021learning} to downstream tasks has emerged as a prominent scheme in centralized learning. In particular, Transformer~\cite{vaswani2017attention,dosovitskiy2020vit} architectures have demonstrated exceptional ability in deriving robust and discriminative representations. In the FL community, some works~\cite{qu2022rethinking,nguyen2022begin,chen2022pre} start to investigate the effectiveness of leveraging foundation models into the FL framework. 
For instance, ViT-FL~\cite{qu2022rethinking} first incorporates the pre-trained Vision Transformer (ViT)~\cite{dosovitskiy2020vit} architecture into FL and shows improved model performance and training stability.
However, most FL algorithms typically require updating \textit{entire} model, making the adoption of foundation models challenging in real-world FL scenarios (\textit{e.g.,} edge devices or medical sites) due to limited computation/communication resources.

Prompt learning techniques~\cite{li2021prefix, liu2021p, lester2021power} have been widely used in the NLP community for adapting language models to downstream tasks effectively via only optimizing a small amount of continuous task-specific prompt vectors. Recently, Visual Prompt Tuning (VPT)~\cite{jia2022visual} has also been proposed as an efficient and effective alternative to fully fine-tuning the large-scale ViT model. It introduces additional learnable prompts into the input image embedding space. These prompts act as task-specific parameters, adapting the frozen backbone model to perform downstream tasks. 
Very recently, several concurrent works~\cite{guo2022promptfl,su2022cross} choose to insert prompts to a frozen CLIP~\cite{radford2021learning} text encoder at local clients.
While allowing efficient FL, these methods follow FedAvg and adopt \textit{average-based} prompt aggregation, which is not optimal for clients with significant data heterogeneity.
Thus, applying prompt learning techniques to data heterogeneous FL scenarios remains an open research challenge. In this work, we propose a unique personalized prompt generation to enable efficient model personalization upon clients with heterogeneous data.

\section{Proposed Method}
\label{sec:method}

\subsection{Problem Formulation} \label{problem}

For the sake of completeness, we first define the problem setting in this paper. Following previous personalized federated learning works~\cite{fallah2020personalized,collins2021exploiting,li2021ditto,shamsian2021personalized,ma2022layer,chen2021bridging,oh2022fedbabu,zhang2022fedala}, we assume that training data are distributed in $N$ separated clients with heterogeneous datasets $\mathcal{D}=\{\mathcal{D}_1,\mathcal{D}_2,...,\mathcal{D}_N\}$, each contains a set of image-label pairs $\mathcal{D}_n=\{(\mathtt{\textbf{x}}_i, y_{i})\}_{i=1}^{|\mathcal{D}_n|}$. These datasets follow \textit{non-IID} (independent and identically distributed) data distribution in terms of either domain discrepancy or imbalanced label space.
With the interest of training efficiency and local data privacy preserved, we aim at learning a client-specific prompt generation mechanism that produces $K$ personalized visual prompts $\textbf{P}_n=[p^1_n, p^2_n,...,p^K_n]$ that adapt a pre-trained foundation model $F^*$ to perform classification tasks on each local client.
Through our learned client-specific prompts, we enable efficient model personalization for each heterogeneous client while preserving the robust representation from a frozen foundation model without the risks of overfitting.

\subsection{Efficient Model Personalization in FL via Client-Specific Prompt Generation} \label{pFedPG}

As illustrated in Fig.~\ref{fig:archi}, we propose a personalized federated learning framework of \emph{client-specific Prompt Generation} (pFedPG). 
To leverage underlying client characteristics and enable efficient model personalization for all clients, pFedPG alternates between the stages of \emph{personalized prompt adaptation} and \emph{personalized prompt generation} at local clients and the global server, respectively.

In the stage of \emph{personalized prompt adaptation}, pFedPG advances the visual prompt learning technique~\cite{jia2022visual} in FL frameworks. A small number of trainable parameters, denoted as \textit{prompts} $\textbf{P}_n=[p^1_n, p^2_n,...,p^K_n]$, are inserted into a frozen foundation model $F^*$ to encode client-specific information at client $n$.
In the stage of \emph{personalized prompt generation}, a personalized prompt generator $G$ is learned to produce personalized prompts for each client by exploiting the underlying characteristics among clients.
Once the learning process is complete, we are able to efficiently adapt the frozen foundation model $F^*$ by the client-specific prompts $\textbf{P}_n$ to perform recognition tasks at each client $n$. We now detail each learning stage, including the training/inference processes below.

\subsubsection{Personalized prompt adaptation at local clients} \label{client}

To enable efficient model adaptation on top of large-scale foundation models and prevent possible overfitting problems caused by updating on relatively limited private data, we advance \emph{Personalized Prompt Adaptation} based on the prompt learning~\cite{jia2022visual} scheme. Note that, the prompts could be treated as client-specific learnable parameters and directly optimized through gradients during training.
With the prompts learned, we can efficiently adapt the foundation model $F^*$ to the data distribution of interest.

As depicted in Fig.~\ref{fig:archi}(a), this training stage aims to learn client-specific prompts $\textbf{P}_n=[p^1_n, p^2_n,...,p^K_n]$ by leveraging the Transformer-based frozen foundation model $F^*$ with locally updated classification head $H_n$. To be more specific, we follow~\cite{dosovitskiy2020vit} and divide an input image $\mathtt{\textbf{x}}$ to $m$ image patches $\{a^i\}_{i=1}^m$ and then derive the latent embedding $\mathtt{\textbf{z}}$ by a frozen feature embedding module $\mathtt{Embed}$ as follows:
\begin{equation}
\begin{aligned}
\mathtt{\textbf{x}}&=[a^1,a^2,...,a^m], \quad a \in \mathbb{R}^{3 \times h \times w}, \\
\mathtt{\textbf{z}}&=[z^1,z^2,...,z^m], \quad z=\mathtt{Embed}(a), \\
\end{aligned}
\label{eq:embed}
\end{equation}
\noindent where $h$ and $w$ denote the height and width of an image patch, and the patch embedding $z^m$ is projected to $l$-dimension. Once the latent embedding $\mathtt{\textbf{z}}$ is obtained, we form the input embedding of the Transformer encoder $F^*$ by concatenating $\mathtt{\textbf{z}}$ with a classification token $c \in \mathbb{R}^l$ (pre-trained with the ViT backbone), and the client-specific prompts $\textbf{P}_n=\left[p^1_n, p^2_n,...,p^K_n\right]$ as $[c, \textbf{P}_n, \mathtt{\textbf{z}}]$. To encourage the client-specific prompts to adapt upon this client's data, we employ the standard cross-entropy loss $\mathcal{L}_{cla}$ over $|\mathcal{D}_n|$ samples, and is calculated as:
\begin{equation}
\begin{aligned}
\mathcal{L}_{n}=\frac{1}{\left| \mathcal{D}_n \right|}\sum_{j=1}^{\left| \mathcal{D}_n \right|}\mathcal{L}_{cla}\left(H_n\left(F^*\left(\left[c,\textbf{P}_n, \mathtt{\textbf{z}}_{j}\right]\right)\right), y_{j}\right).
\end{aligned}
\label{eq:ce}
\end{equation}

\noindent As a result, the client-specific prompts $\mathtt{\textbf{P}_n}$ can be optimized end-to-end by gradient decent (the same as $H_n$) with learning rate $\gamma$ as $\mathtt{\widetilde{\textbf{P}}_n}\leftarrow \mathtt{\textbf{P}_n} - \gamma \cdot \partial (\mathcal{L}_{n}) /\partial \mathtt{\textbf{P}_n}$.

With personalized prompt adaptation, pFedPG is able to realize parameter-efficient model adaptation without requiring updating entire model parameters yet mitigating possible overfitting concerns and huge computation workloads.

\subsubsection{Personalized prompt generation at the server} \label{server}

Conventional FL methods (\textit{e.g.,} \cite{mcmahan2017communication}) typically adopt average-based model aggregation at the server. However, this aggregation manner poses a significant risk of deviating from local data distributions and introduces massive communication overheads, especially when deploying large-scale models among heterogeneous clients. Recall that the prompts trained locally could be treated as client-specific parameters to adapt the frozen model to the client of interest. Instead of averaging model parameters or prompts from clients, we aim at learning a unique personalized prompt generation mechanism at the server to exploit cross-client knowledge and then produce personalized prompts that serve as a good initialization to facilitate efficient local adaptation. Since the server cannot access local private data, it is challenging to obtain the client-specific characteristics for encouraging the produced personalized prompts to boost local adaptation. In the following, we will elaborate on how our personalized prompt generation be learned in the FL scheme.



\paragraph{Design and architecture}
As illustrated in Fig.~\ref{fig:archi}(b), with the goal of generating personalized prompts $\{\textbf{P}_1,...\textbf{P}_N\}$ for all $N$ clients, our pFedPG learns to transform a set of client-agnostic prompt basis $\textbf{P}_{base}$ through a conditional prompt generator $G(\cdot;\varphi)$ parameterized by $\varphi$ with the guidance of client descriptor $d_n$ selected from $D=\{d_1,d_2,...,d_N\}$. 
To be more specific, we realize the conditional prompt generator $G$ based on cross-attention~\cite{vaswani2017attention} while the client-agnostic prompts $\textbf{P}_{base}$ and the client descriptor $d_n$ are expected to capture client-agnostic information and encode the client-specific characteristics, respectively.
As a result, generating personalized prompts could be achieved by retrieving client-relevant knowledge from $\textbf{P}_{base}$ through the query of the client descriptor $d_n$, as formulated below,
\begin{equation}
\begin{aligned}
\textbf{P}_n&=G\left( \textbf{P}_{base},d_n \right) =\textbf{P}_{base}+\mathtt{Atten}\left( \mathcal{Q},\mathcal{K},\mathcal{V} \right)W^\mathcal{O}\\
&=\textbf{P}_{base}+\mathtt{Softmax}(\frac{\mathcal{Q}\mathcal{K}^{T}}{\sqrt{l_k}})\mathcal{V}W^\mathcal{O},
\end{aligned}
\label{eq:atten}
\end{equation}
\begin{equation*}
\begin{aligned}
\text{where} \quad \mathcal{Q}=[d_n]W^{\mathcal{Q}}, \mathcal{K}=\textbf{P}_{base}W^{\mathcal{K}},\mathcal{V}=\textbf{P}_{base}W^{\mathcal{V}},
\end{aligned}
\label{eq:qkv}
\end{equation*}
\noindent where $\sqrt{l_k}$ is a scaling factor and $l$ is the embedding dimension. $W^{\mathcal{Q}}\in \mathbb{R}^{l\times l_k}$, $W^{\mathcal{K}}\in \mathbb{R}^{l\times l_k}$, $W^{\mathcal{V}}\in \mathbb{R}^{l\times l_v}$, and $W^{\mathcal{O}}\in \mathbb{R}^{l_v\times l}$ are learnable projection matrixes, where $l_k$ and $l_v$ are internal dimensions, as in \cite{vaswani2017attention}.

\paragraph{Learning of personalized prompt generation}
As the goal of personalized prompts is to serve as a good initialization for each client that facilitates the local adaptation, we learn our personalized prompt generation module (\textit{i.e.,} $G$, $\textbf{P}_{base}$ and $d_n$) through the training rewards observed from the local optimization process.
Inspired by~\cite{shamsian2021personalized,ma2022layer}, the change of prompts after local training $\Delta \textbf{P}_n=\widetilde{\textbf{P}_n}-\textbf{P}_n$ indicates the direction of local optimization at client $n$ that could be treated as training feedback, assessing the quality of the server-generated prompt initialization for each client.
With $\Delta \textbf{P}_n$ observed, we are capable of training our pFedPG end-to-end via gradient descent.

To be more specific, the update of the conditional prompt generator $G(\cdot;\varphi)$ can be derived by the gradients computed locally and expressed by the chain rule as
\begin{equation}
\begin{aligned}
\Delta\varphi=\nabla_\varphi\mathcal{L}_n&=(\nabla_\varphi\textbf{P}_n)^T\nabla_{\textbf{P}_n}\mathcal{L}_n \\&\cong (\nabla_\varphi\textbf{P}_n)^T\Delta \textbf{P}_n,  
\end{aligned}
\label{eq:delta}
\end{equation}
\noindent where $\nabla_{\textbf{P}_n}\mathcal{L}_n$ is approximated by $\Delta \textbf{P}_n$ that indicates the optimization direction of local training. We apply the same optimization rule to learn the client-agnostic prompts $\textbf{P}_{base}$ and client descriptor $d_n$ end-to-end with $G$, and summarize the gradient update as follows,
\begin{equation}
\begin{aligned}
\varphi\gets \varphi-\alpha \nabla_{\varphi}\textbf{P}_n^T\Delta \textbf{P}_n,\\
\textbf{P}_{base}\gets \textbf{P}_{base}-\alpha\nabla_{\textbf{P}_{base}}\varphi^T \nabla_{\varphi}\textbf{P}_n^T\Delta \textbf{P}_n, \\
d_n\gets d_n-\alpha\nabla_{d_n}\varphi^T \nabla_{\varphi}\textbf{P}_n^T\Delta \textbf{P}_n.
\end{aligned}
\label{eq:pg}
\end{equation}

\begin{algorithm}[tb]
\caption{pFedPG for Efficient and Personalized FL}
\label{alg:algorithm}
\textbf{Input}: Number of communication rounds $T$, $F^*$, $G$, $\textbf{P}_{base}$, $D$, and $N$ sets of $\textbf{P}_n$ and $H_n$, $n \in [1, N]$ \\
\textbf{Data}: $N$ labeled datasets $\mathcal{D}_n$, $n \in [1, N]$\\
\textbf{Output}: $F^*$, $H_n$, $\textbf{P}_n$
\begin{algorithmic}[1] 
\STATE Let $t=0$;
\WHILE{$t$ \textless $T$}
\STATE \textbf{\# Personalized prompt adaptation at clients}
\FOR{$n$ in $1:N$}
\STATE Keep $F^{*}$ freeze;
\STATE Set $\textbf{P}_n=G(\textbf{P}_{base},d_n)$, $d_n \in D$ (Eq.~\eqref{eq:atten});
\STATE Randomly sample a minibatch from $\mathcal{D}_n$;
\STATE Update $H_n$ with $\mathcal{L}_{n}$ (Eq.~\eqref{eq:ce});
\STATE Update $\textbf{P}_n$ by $\mathtt{\widetilde{\textbf{P}}_n}\leftarrow \mathtt{\textbf{P}_n} - \gamma \frac{\partial (\mathcal{L}_{n})}{\partial \mathtt{\textbf{P}_n}}$;
\STATE $\Delta \textbf{P}_n=\widetilde{\textbf{P}_n}-\textbf{P}_n$;
\ENDFOR
\STATE \textbf{\# Personalized prompt generation at the server}
\STATE Receive $\Delta \textbf{P}_n$ from all $N$ clients;
\STATE Update $G$, $\textbf{P}_{base}$, and $D$ by Eq.~\eqref{eq:pg};
\STATE $t = t + 1$;
\ENDWHILE
\end{algorithmic}
\end{algorithm}

We note that, the client-agnostic prompt basis $\textbf{P}_{base}$ and conditional prompt generator $G$ are optimized by all clients, enforcing them to exploit cross-client knowledge, while client descriptor $d_n$ is solely regarding client $n$, to encourage the derivation of client-specific characteristics.
With our proposed personalized prompt generation module, pFedPG is able to generate personalized prompts to facilitate local adaptation while leveraging learned knowledge across clients without explicitly accessing private data.

\begin{table*}[tp!]
\caption{Quantitative comparisons on Office-Caltech10 and DomainNet datasets using ViT-B/16. \textbf{Bold} denotes the best result.}
\label{tab:domain}
\centering
\vspace{1mm}
\resizebox{0.97\textwidth}{!}{
\begin{tabular}{@{}lccccccccccccc@{}}
\toprule
Datasets & \multicolumn{5}{c}{Office-Caltech10 (\%)} & \multicolumn{7}{c}{DomainNet (\%)} & Comm. \\ 
\cmidrule(r){1-1}  \cmidrule(lr){2-6} \cmidrule(lr){7-13} 
Method           & \textit{A} & \textit{C} & \textit{D}  & \textit{W} & Avg. & \textit{C}        & \textit{I}      & \textit{P}       & \textit{Q} & \textit{R} & \textit{S} & Avg.  & Cost \\ \midrule
\multicolumn{6}{@{}l@{}}{\textit{\textbf{Baselines}}} \\
SingleSet-Full   & 80.73           & 73.33            & 90.62          & 94.92           & 84.90                 & 47.34          & 37.14          & 67.21          & 55.30     & 84.88     & 45.13       & 56.17        & -                 \\
SingleSet-VPT~\cite{jia2022visual} & 83.33           & 74.67            & 96.88          & 96.61           & 87.87                 & 57.98          & 41.55          & 74.64          & 59.60     & 89.56     & 60.47       & 63.97        & -                 \\ 
FedAvg~\cite{qu2022rethinking}      & 89.58                & 80.44                 & 100.0               & 100.0                & 92.51                 & 63.50               & 38.05               & 71.89               & 60.80         & 78.55     & 60.47       & 62.21             & $8.58 \times 10^{7}$                 \\ \midrule
\multicolumn{6}{@{}l@{}}{\textit{\textbf{Personalized Federated Learning}}} \\
Per-FedAvg~\cite{fallah2020personalized}      & 91.67                & 90.22                 & 100.0               & 100.0                & 95.47                 & 69.39               & 48.71               & 82.07               & 35.30         & 90.63     & 72.56       & 66.44          & $8.58 \times 10^{7}$                   \\
FedRep~\cite{collins2021exploiting}      & 91.15                & 88.44                 & 100.0               & 100.0                & 94.90                 & 64.26               & 38.20               & 72.86               & \textbf{62.10}         & 82.66     & 60.11       & 63.37         & $8.58 \times 10^{7}$                       \\ 
FedRoD~\cite{chen2021bridging}      & 92.19                & 90.67                 & 100.0               &  100.0               & 95.72                 & 66.54               & 42.92               & 74.15               & 57.20         & 84.63     & 66.43       & 65.31          & $8.58 \times 10^{7}$                     \\ 
FedBABU~\cite{oh2022fedbabu}      & 89.06                & 85.78                 & 100.0               &  100.0               & 93.71                 & 63.31               & 43.07               & 74.80               & 43.80         & 87.26     & 67.15       & 63.23           & $8.58 \times 10^{7}$                     \\ \midrule
\multicolumn{6}{@{}l@{}}{\textit{\textbf{Efficient Federated Learning}}} \\
FedVPT~\cite{jia2022visual}    & 92.71           & 84.44            & 100.0          & 100.0           & 94.29                & 65.59          & 44.14          & 76.58          & 47.30     & 91.04     & 60.29       & 64.16       & $7.68 \times 10^{3}$                   \\
FedVPT-D~\cite{jia2022visual}    & 91.67           & 89.33            & 100.0          & 100.0           & 95.25                & 63.31          & 43.07          & 74.80          & 54.80     & 87.26     & 67.15       & 65.07       & $9.22 \times 10^{3}$                   \\
pFedPG (Ours)    & \textbf{94.79}  & \textbf{92.44}   & \textbf{100.0} & \textbf{100.0}           & \textbf{96.81}                & \textbf{73.00}  & \textbf{50.08}  & \textbf{84.33}  & 60.00     & \textbf{94.00}      & \textbf{68.41}        & \textbf{71.64}         & $7.68 \times 10^{3}$           \\ \bottomrule
\end{tabular}
}
\vspace{0mm}
\end{table*}

\subsection{pFedPG Training and Inference} \label{inference}

In Algorithm~\ref{alg:algorithm}, we summarize the training details of our proposed pFedPG. We alternate between the learning processes of personalized prompt generation and personalized prompt adaptation until converging.

Once the learning of the proposed framework is complete, we deploy the learned client-specific prompts $\textbf{P}_n$ to instruct the pre-trained feature extractor $F^*$ to extract discriminative representations together with locally trained classification head $H_n$ for performing the recognition task at each client. Formally, the categorical predictions $y^*$ over $Y$ classes at each client $n$ can be computed as: 
\begin{equation}
\begin{aligned}
y^*=\mathop{\arg\min}_{k \in K}{H_n(F^*(\left[c, \textbf{P}_n,\mathtt{\textbf{x}}\right]))}.
\end{aligned}
\label{eq:pred}
\end{equation}

\section{Experiments}
\label{sec:exp}

\subsection{Datasets and Experimental Setup}
\subsubsection{Datasets} \label{ssec:dataset}
We evaluate our method on five public benchmark datasets covering types of data heterogeneity, including domain discrepancy and imbalanced class distribution. For \textit{domain discrepancy}, \textbf{Office-Caltech10}~\cite{saenko2010adapting,griffin2007caltech} is composed of four data domains including \textit{Amazon}, \textit{DSLR}, \textit{Webcam}, and \textit{Caltech}. Each domain contains ten classes, with 2,533 images in total.
\textbf{DomainNet}~\cite{peng2019moment} consists of 0.6 million images of 345 classes distributed across six domains, \textit{Clipart}, \textit{Infograph}, \textit{Painting}, \textit{Quickdraw}, \textit{Real} and \textit{Sketch}. Following~\cite{li2021fedbn}, we use the top ten most frequent classes to form a sub-dataset for our experiments.
As for medical image diagnosis tasks, \textbf{Dermoscopic-FL}~\cite{chen2021personalized} is comprised of four data sites collected from HAM10K~\cite{tschandl2018ham10000} and MSK~\cite{codella2018skin}. Each data site contains three types of skin lesions, with 10,490 images in total. More detailed statistics and sampled images are provided in the supplementary material.
For \textit{imbalanced class distribution}, \textbf{CIFAR-10}~\cite{krizhevsky2009learning} contains 5,000 training images and 1,000 testing images per class, totaling ten classes. \textbf{CIFAR-100}~\cite{krizhevsky2009learning} consists of 60,000 images of 100 categories with 500 training images and 100 testing images per class. 

\subsubsection{Experimental settings} \label{ssec:exp_setting}
To properly evaluate our proposed approach and fairly compare it with existing FL methods, we conduct experiments on two types of heterogeneous FL settings: domain discrepancy and imbalanced class distribution.
For conducting clients with \textit{domain discrepancy}, we assign a data domain to a client, indicating the number of clients ($N$) is set as 4, 6, and 4 for Office-Caltech10, DomainNet, and Dermoscopic-FL datasets, respectively.
As for simulating \textit{imbalanced class distribution}, we consider two non-IID settings using CIFAR-10 and CIFAR-100. Following~\cite{qu2022rethinking}, the first non-IID setting we considered is randomly selecting disjoint $c$ classes for each client and denoted as \emph{disjoint label space}. In our experiments, $c=2$ and $c=10$ for CIFAR-10 and CIFAR-100, respectively.
As for the other non-IID setting, data in each class would be partitioned into all clients following a Dirichlet distribution $Dir(\alpha)$. We follow \cite{chen2021bridging} and set $\alpha$ to 0.1 over 10 clients.

\subsubsection{Implementation details} \label{ssec:implement}

We use ViT-B/16~\cite{dosovitskiy2020vit} pre-trained on ImageNet21k~\cite{deng2009imagenet} as the backbone of $F^*$ and a single linear layer to realize the classification head $H_n$. The input images of all datasets are resized to 224 $\times$ 224 pixels. For each client, we train $\mathtt{\textbf{P}_n}$ and $H_n$ using the SGD optimizer with a learning rate $\gamma$ of 0.25 with a weight decay rate of 0.001 and a batch size of 64 for 5 epochs. The number of communication round $T$ is set to 100. We set the learning rate $\alpha$ for updating $G$, $\textbf{P}_{base}$, and $D$ to 0.001. The number of prompts $K$ of $\mathtt{\textbf{P}_n}$ and $\textbf{P}_{base}$ is set as 10 for datasets except for Dermoscopic-FL with $K=3$. The hyperparameters above are tuned by cross-validation.
In all our experiments, we implement our model using PyTorch~\cite{paszke2019pytorch} and conduct training on NVIDIA TESLA V100 GPUs with 32 GB memory.

\begin{table}[tp!]
\caption{Quantitative comparisons on CIFAR-10/100 datasets using ViT-B/16. \textbf{Bold} denotes the best result.}
\label{tab:class}
\centering
\vspace{1mm}
\resizebox{0.45\textwidth}{!}{
\begin{tabular}{@{}lcccc}
\toprule
Datasets & \multicolumn{2}{c}{CIFAR-10 (\%)} & \multicolumn{2}{c}{CIFAR-100 (\%)} \\ \cmidrule(r){1-1}  \cmidrule(lr){2-3} \cmidrule(lr){4-5}
Method & Disjoint & $Dir(0.1)$ & Disjoint & $Dir(0.1)$ \\ \midrule
\multicolumn{5}{@{}l@{}}{\textit{\textbf{Baselines}}} \\
SingleSet-Full & 89.51 & 83.85 & 67.74 & 49.64 \\
SingleSet-VPT~\cite{jia2022visual} & 88.91 & 84.32 & 63.42 & 46.46 \\ 
FedAvg~\cite{qu2022rethinking} & 88.04 & 79.79 & 63.33 & 51.37 \\ \midrule
\multicolumn{5}{@{}l@{}}{\textit{\textbf{Personalized Federated Learning}}} \\
Per-FedAvg~\cite{fallah2020personalized} & 88.13  & 85.14  & 69.31  & 52.68  \\
FedRep~\cite{collins2021exploiting} & 87.07 & 82.40 & 65.71 & 50.36 \\
FedRoD~\cite{chen2021bridging} & 87.61 & 80.36 & 63.90 & 51.42 \\
FedBABU~\cite{oh2022fedbabu} & 83.15  & 76.33 & 55.91  & 50.19  \\ \midrule
\multicolumn{5}{@{}l@{}}{\textit{\textbf{Efficient Federated Learning}}} \\
FedVPT~\cite{jia2022visual} & 89.39 & 85.11 & 55.49 & 45.26 \\
FedVPT-D~\cite{jia2022visual} & 89.56 & 85.43 & 66.91 & 50.25 \\
pFedPG (Ours) & \textbf{90.08} & \textbf{87.57} & \textbf{70.96} & \textbf{55.91} \\ \bottomrule
\end{tabular}
}
\vspace{0mm}
\end{table}

\subsection{Quantitative Evaluation} \label{ssec:quan}

We compare our proposed pFedPG with existing FL methods on benchmark datasets representing various types of data heterogeneity (\textit{i.e.,} domain discrepancy and imbalanced class distribution). In our experiments, \textit{SingleSet-Full} and FedAvg~\cite{mcmahan2017communication} are viewed as baselines, where the former trains a model at each client without information sharing, while the latter aggregates client models to construct a shared global model. In addition, \textit{SingleSet-VPT} indicates each client independently applies visual prompt tuning~\cite{jia2022visual} to learn prompts at the input embedding space.

In Tables~\ref{tab:domain}-\ref{tab:medical}, we summarized the results compared with the state-of-the-art pFL works. To be more specific, Per-FedAvg~\cite{fallah2020personalized} applies meta-learning~\cite{finn2017model} to derive customized models for each client from a global initialization.
FedRep~\cite{collins2021exploiting} aggregates feature extractors but keeps classifiers trained locally; FedBABU~\cite{oh2022fedbabu} only updates and shares feature extractors during FL training.
FedRoD~\cite{chen2021bridging} additionally learns a personalized classification head without model aggregation.
Instead of updating entire model parameters, two \textit{efficient} FL baselines, \textit{FedVPT} and \textit{FedVPT-D}, are conducted, which keep the backbone frozen, and aggregate prompts globally. Following~\cite{jia2022visual}, FedVPT inserts prompts to the input, and FedVPT-D prepends prompts to the input and hidden layers.
Note that, we use ViT-B/16~\cite{dosovitskiy2020vit} as the backbone of the above methods for fair comparisons.

In Table~\ref{tab:domain}, we provide the quantitative comparisons on Office-Caltech10 and DomainNet datasets with the presence of \textbf{domain shifts} across clients. Our approach achieved the highest 96.81\% and 71.64\% average accuracies on Office-Caltech10 and DomainNet, respectively, as shown from Table~\ref{tab:domain}. Furthermore, our method demonstrated the best communication efficiency, using only approximately 0.01\% of parameters in comparison to other existing pFL methods. Note that the costs of FedRep~\cite{collins2021exploiting} and FedBABU~\cite{oh2022fedbabu} are the numbers of model parameters of the ViT backbone (\textit{i.e.,} 85.8M), while the communication costs of~\cite{mcmahan2017communication,fallah2020personalized,chen2021bridging} can be approximated to 85.8M, as they transmit the ViT backbone along with a single-layer classifier, which adds relatively few parameters.

In addition to domain discrepancy, we conducted comparisons on the \textbf{imbalanced class distribution} scenario using CIFAR-10 and CIFAR-100 datasets, as shown in Table~\ref{tab:class}. 
As mentioned in Sec.~\ref{ssec:exp_setting}, two types of imbalanced data are simulated, including disjoint label space and imbalanced label distribution drawn from $Dir(0.1)$. 
Table~\ref{tab:class} demonstrates that our method performed favorably against existing FL works over the two datasets on both types of label imbalance.
To further exhibit the ability of our method to more practical scenarios, we compare with state-of-the-art works for the cross-site medical image diagnosis task using Dermoscopic-FL. As we can observe in Table~\ref{tab:medical}, our pFedPG consistently performed superiorly against other FL methods on all hospital sites.

\begin{table}[tp!]
\caption{Quantitative comparisons on Dermoscopic-FL dataset using ViT-B/16. \textbf{Bold} denotes the best result.}
\label{tab:medical}
\centering
\vspace{1mm}
\resizebox{0.45\textwidth}{!}{
\begin{tabular}{@{}lccccc}
\toprule
Method           & A     & B     & C     & D     & Avg. \\ \midrule
\multicolumn{5}{@{}l@{}}{\textit{\textbf{Baselines}}} \\
SingleSet-Full   & 76.09      & 97.29      & 71.65      & 73.57      & 79.65        \\
SingleSet-VPT~\cite{jia2022visual} & 70.90      & 96.25      & 70.12      & 68.33      & 76.40        \\ 
FedAvg~\cite{qu2022rethinking}      & 62.54 & 96.12 & 51.52 & 68.08 & 69.57   \\ \midrule
\multicolumn{5}{@{}l@{}}{\textit{\textbf{Personalized Federated Learning}}} \\
Per-FedAvg~\cite{fallah2020personalized}      & 76.09 & 91.99 & 70.12 & 74.56 & 78.19   \\
FedRep~\cite{collins2021exploiting}      & 69.06 & 96.12 & 60.37 & 68.58 & 73.53   \\
FedRoD~\cite{chen2021bridging}      & 63.55 & 96.67 & 58.84 & 69.33 & 72.10   \\
FedBABU~\cite{oh2022fedbabu}      & 58.19 & 97.16 & 49.09 & 68.58 & 68.26   \\ \midrule
\multicolumn{5}{@{}l@{}}{\textit{\textbf{Efficient Federated Learning}}} \\
FedVPT~\cite{jia2022visual}    & 74.92 & 96.77 & 67.07 & 75.06 & 78.46   \\
FedVPT-D~\cite{jia2022visual}    & 73.91 & 96.12  & 74.09  & 77.81  & 80.48    \\
pFedPG (Ours)             & \textbf{79.26} & \textbf{97.29} & \textbf{76.22} & \textbf{78.80} & \textbf{82.89}   \\ \bottomrule
\end{tabular}
}
\vspace{0mm}
\end{table}

\begin{table*}[tp!]
\caption{Analysis of our personalized prompt generation and the architecture of prompt generator $G$ on benchmark datasets.}
\label{tab:aba}
\centering
\vspace{1mm}
\resizebox{0.72\textwidth}{!}{
\begin{tabular}{@{}lccccc}
\toprule
Module & Method & Office-Caltech10 & DomainNet & CIFAR-10 & CIFAR-100 \\
\midrule
 \multirow{2}{*}{Prompt generation} & FedVPT & 94.29  & 64.16  & 89.39  & 55.49  \\
 &  $\textbf{P}_{base}$ & 93.16  & 64.87  & 88.23  & 66.89  \\ 
\midrule 
\multirow{2}{*}{Architecture of $G$} & MLP~\cite{shamsian2021personalized} & 94.96 & 63.33 & 87.47 & 66.73 \\
 & AdaIN~\cite{huang2017arbitrary} & 95.72 & 70.08 & 89.77  & 69.44  \\ \midrule
  & \textbf{pFedPG}   & \textbf{96.81} & \textbf{71.64} & \textbf{90.08} & \textbf{70.96} \\ \bottomrule
\end{tabular}
}
\vspace{0mm}
\end{table*}

We observed that, with the presence of significant data heterogeneity (\textit{e.g.,} large style difference in DomainNet) across clients, existing FL works which obtain a shared feature encoder~\cite{collins2021exploiting,chen2021bridging,oh2022fedbabu} by aggregation might still deviate from local data domains, while Per-FedAvg~\cite{fallah2020personalized} focuses on deriving a global initialization would not be preferable under severe discrepancy across clients. As shown in Tables~\ref{tab:domain}-~\ref{tab:medical}, FedVPT and FedVPT-D achieve comparable or even superior performance over existing FL works, exhibiting the ability of efficient FL methods to mitigate possible overfitting issues. However, sharing a set of global prompts is still not desirable for heterogeneous clients.
To explicitly enable efficient model personalization to tackle heterogeneous data, our approach learns to generate personalized prompts to facilitate local adaptation for each client. With the above results, we successfully confirm the effectiveness and robustness of our proposed pFedPG to address data heterogeneity with training efficiency.



\subsection{Analysis of Our pFedPG} \label{ssec:ablation}

In this section, we first conduct experiments to confirm the effectiveness of our designed personalized prompt generation. Then, we provide a detailed analysis of the impact of different number prompts.
Due to the page limitations, we provide the analysis of model backbones and the size of client data in the supplementary material.

\paragraph{Effectiveness of personalized prompt generation}
In the upper part of Table~\ref{tab:aba}, we intend to verify the effectiveness of our personalized prompt generation for facilitating adaptation at each client on benchmark datasets, where CIFAR-10/100 are under the setting of disjoint label space. In Table~\ref{tab:aba}, we first ablate $\textbf{P}_n$ with the global prompts obtained by global averaging (as in \textit{FedVPT}). 
As reported in Table~\ref{tab:aba}, the globally averaged prompts cannot achieve satisfactory performance since sharing a single set of prompts would not be favorable to heterogeneous clients.
In addition, we examine the performance of applying the trained \textit{client-agnostic prompt basis} $\textbf{P}_{base}$ to clients instead of applying personalized prompts $\textbf{P}_n$.
We observed that the performance of $\textbf{P}_{base}$ is still inferior to ours (which applies $\textbf{P}_n$).
As evident from the above experiments, the effectiveness of our proposed personalized prompt generation for allowing personalized FL under various types of data heterogeneity would be successfully verified.

\paragraph{Effectiveness of our designed prompt generator $G$}
From the results shown in the lower half of Table~\ref{tab:aba}, we see that the performance dropped when we replaced our cross-attention-based prompt generator $G$ and $\textbf{P}_{base}$ with an MLP-based network as~\cite{shamsian2021personalized}, which acts on client descriptors and then output prompts for each client. 
The inferior performance of the MLP-based prompt generator is due to its high training complexity and instability, resulting from the requirement of deploying a fully-connected layer for each prompt embedding. Another alternative prompt generator is to compute adaptive instance normalization (AdaIN)~\cite{huang2017arbitrary} for $\textbf{P}_{base}$ and the client descriptor $d_n$. This method allows for the transfer of client-agnostic prompts $\textbf{P}_{base}$ to personalized prompts $\textbf{P}_n$ by replacing the mean and variance calculated from the client descriptor $d_n$, similar to the style transfer approach~\cite{huang2017arbitrary}. However, as seen in Table~\ref{tab:aba}, directly computing AdaIN did not explicitly model the prompt generation process, resulting in inferior performance compared to ours.
The results summarized in Table~\ref{tab:aba} confirm the effectiveness of our designed architecture of prompt generator $G$.

\paragraph{Impact of the number of prompts $K$}
We also analyze the impact of the number of prompts $K$ on benchmark datasets, and show the results in Table~\ref{tab:num_pt}. 
We found that when the number of prompts is set too low (\textit{e.g.,} $K=1$), the model's accuracy drops slightly due to insufficient capacity. In contrast, if the number of prompts is set too high, such as 100 or 200, the model's performance significantly degrades. This is because a large number of prompts may encode noisy and task-irrelevant information, which can adversely affect the quality of the features derived from foundation models.
With the above observation, we thus set $K$ as 10 for these datasets which achieves the best trade-off between communication cost and performance.


\begin{table}[tp!]
\caption{Impact of the number of prompts $K$ on benchmark datasets, where CIFAR-10/100 are drawn from $Dir(0.1)$.}
\label{tab:num_pt}
\centering
\vspace{1mm}
\resizebox{0.46\textwidth}{!}{
\begin{tabular}{@{}lcccc}
\toprule
$K$ & Office-Caltech10  & DomainNet & CIFAR-10 & CIFAR-100  \\
\midrule
1 & 96.09 & 70.27 & 86.14  & 55.77 \\
5 & 96.77 & 70.53 & 87.41  & 55.79 \\
10 & \textbf{96.81} & \textbf{71.64} & \textbf{87.57}   & \textbf{55.91} \\
50 & 95.10 & 69.55 & 85.63  & 54.52 \\
100 & 94.53 & 68.79 & 85.02  & 53.61 \\
200 & 94.46 & 66.83 & 83.53 & 52.34 \\ \bottomrule
\end{tabular}
}
\vspace{0mm}
\end{table}


\section{Conclusion}
\label{sec:con}

In this paper, we proposed a novel client-specific Prompt Generation framework (pFedPG) for enabling efficient model personalization among heterogeneous clients.
By alternative optimization of the proposed personalized prompt generation and client-specific prompt adaptation, our pFedPG is capable of producing personalized prompts for each client by observing underlying directions of local training among clients, while clients optimize such client-specific prompts to adapt a pre-trained model to local data distribution.
We conducted extensive quantitative experiments, verifying that our framework performed favorably against SOTA pFL approaches at heterogeneous data clients while achieving training and communication efficiency.

\paragraph{Acknowledgment}
This work is supported in part by the National Science and Technology Council under grant NSTC111-2634-F-002-020 and National Taiwan University under grant NTU-112L900901. We also thank to National Center for High-performance Computing (NCHC) for providing computational and storage resources.

{\small
\bibliographystyle{ieee_fullname}
\bibliography{main}
}

\end{document}